\documentclass[letterpaper, 10 pt, conference]{ieeeconf}  

\IEEEoverridecommandlockouts                              

\usepackage{url} 
\usepackage[section]{placeins}
\usepackage{graphicx} 
\usepackage{tikz}
\usepackage{amsmath}
\usepackage{amssymb}
\usepackage{booktabs}
\usepackage{xcolor}
\usepackage{hyperref}
\usepackage{multirow}

\title{\LARGE \bf
BarlowWalk: Self-supervised Representation Learning for Legged Robot Terrain-adaptive Locomotion
}

\author{Haodong Huang$^{1}$, Shilong Sun$^{1,2,3}$, Yuanpeng Wang$^{1}$, Chiyao Li$^{1}$, Hailin Huang$^{1,2,3}$, Wenfu Xu$^{1,2,3}$
\thanks{}
\thanks{$^{1}$Haodong Huang, Shilong Sun, Yuanpeng Wang, Chiyao Li, Hailin Huang and Wenfu Xu are  with School of Robotics and Advanced Manufacture, Harbin Institute of Technology, Shenzhen, 518052, China (\textit{Corresponding author: Shilong Sun})  
        {\tt\small \url{hhd1340201839@163.com}, \url{sunshilong@hit.edu.cn}, \url{qq1226171844@126.com}, \url{lcy_51k@163.com},  \url{huanghailin@hit.edu.cn}, \url{wfxu@hit.edu.cn}}.}%
\thanks{$^{2}$Guangdong Provincial Key Laboratory of Intelligent Morphing Mechanisms and Adaptive Robots,  Shenzhen 518052, China.}%
\thanks{$^{3}$Guangdong Biomimetic Intelligent Unmanned System Engineering Technology Research Center,  Shenzhen 518052, China.}%
}

\begin{document}

\maketitle
\thispagestyle{empty}
\pagestyle{empty}

\begin{abstract}

Reinforcement learning (RL), driven by data-driven methods, has become an effective solution for robot leg motion control problems. However, the mainstream RL methods for bipedal robot terrain traversal, such as teacher-student policy knowledge distillation, suffer from long training times, which limit development efficiency. To address this issue, this paper proposes BarlowWalk, an improved Proximal Policy Optimization (PPO) method integrated with self-supervised representation learning. This method employs the Barlow Twins algorithm to construct a decoupled latent space, mapping historical observation sequences into low-dimensional representations and implementing self-supervision. Meanwhile, the actor requires only proprioceptive information to achieve self-supervised learning over continuous time steps, significantly reducing the dependence on external terrain perception. Simulation experiments demonstrate that this method has significant advantages in complex terrain scenarios. To enhance the credibility of the evaluation, this study compares BarlowWalk with advanced algorithms through comparative tests, and the experimental results verify the effectiveness of the proposed method.

\end{abstract}

\section{Introduction}

Humans exhibit remarkable adaptability to complex terrains through their bipedal structure, yet achieving comparable gait control in legged robots remains a significant challenge. Traditional model-based approaches have yielded promising results by leveraging kinematic and dynamic principles \cite{1}, \cite{2}, \cite{3}, \cite{4}, \cite{5}, \cite{6}, \cite{7}, \cite{8}. However, these methods often struggle to balance model accuracy with computational efficiency, particularly in real-time control scenarios. Additionally, they require an in-depth understanding of the robot’s dynamics, which not only raises the technical barrier for entry but also limits broader participation from researchers with less domain expertise. n recent years, reinforcement learning (RL)-based methods have emerged as a prominent research direction. By leveraging simulation environments and task-specific reward functions, RL enables robots to autonomously acquire complex motor skills without the need for explicit modeling \cite{9}, \cite{10}, \cite{11}, \cite{12}, \cite{13}, \cite{15}, offering new avenues for real-time, adaptive locomotion control.

\begin{figure}
    \centering
    \includegraphics[width=0.9\linewidth]{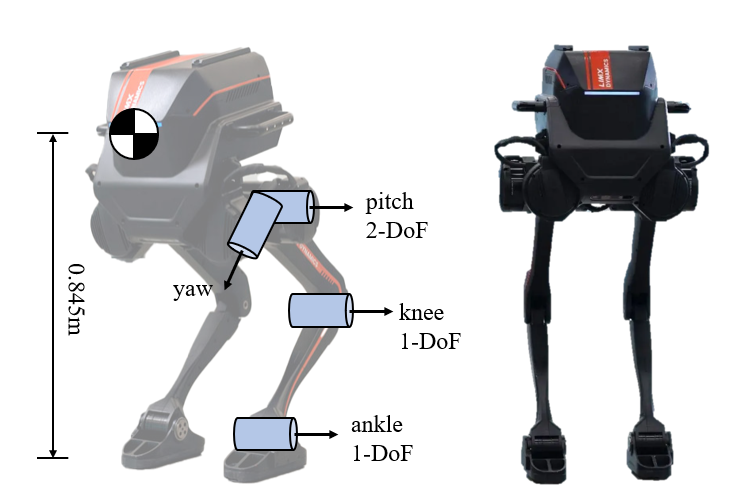}
    \caption{The robot used in the experiment is the TRON1 from Limx Dynamics.}
    \label{fig:robot}
\end{figure}

Current mainstream RL approaches for bipedal robots largely rely on privileged learning methods \cite{17}. In real-world scenarios, however, robots typically lack access to explicit terrain information and must make decisions based solely on proprioceptive inputs. As such, the interaction is commonly modeled as a Partially Observable Markov Decision Process (POMDP).
To address partial observability, existing studies often: (1) feed terrain information into the RL critic network, enabling it to implicitly guide the policy by learning terrain-aware value estimation; or (2) adopt teacher-student frameworks, where a teacher policy with terrain access transfers knowledge to a student policy that relies only on proprioception \cite{18}. Many works leverage the full observability in simulation to facilitate training under these paradigms.

In quadruped robot research, the inherent stability provided by four-legged structures allows some studies to achieve locomotion on complex terrains using only current and recent proprioceptive information as input \cite{19}. In contrast, bipedal robots, due to their structurally limited support, face greater challenges in maintaining balance and stability. As a result, relying solely on proprioceptive inputs makes it significantly more difficult for bipedal robots to walk stably on complex terrains.

In this study, we propose an locomotion control framework for bipedal robots. Specifically, terrain information is provided to the critic network, while the policy network receives only historical proprioceptive data. To enhance the representation of temporal observations, we incorporate a self-supervised learning method, referred to as BarlowWalk, which is based on the Barlow Twins algorithm. This method encodes latent features from fixed-length proprioceptive histories and reduces redundancy to extract more compact and informative representations. Such representations enable the policy to more effectively capture essential motion dynamics, thereby improving the robot’s locomotion performance.We contend that autonomously learned latent representations are inherently more adaptive and expressive than handcrafted features, facilitating better generalization to complex and dynamic environments. Importantly, the proposed approach achieves stable locomotion across diverse terrains without relying on knowledge distillation techniques.

The main contributions of this study can be summarized as follows:

(1) A novel locomotion control framework is proposed for bipedal robots by integrating Barlow Twins self-supervised learning with reinforcement learning.

(2) The policy network is trained solely on historical proprioceptive data, without teacher-student guidance, allowing the robot to learn locomotion strategies from its own experience on complex terrains.

(3) The proposed method is validated on bipedal robots through experiments on complex terrains, demonstrating robust and adaptive locomotion performance..

The structure of this paper is as follows: Section \ref{sec:relatedwork} provides a brief overview of the relevant research progress in the field of deep reinforcement learning. Section \ref{sec:method} details the proposed control framework, including the design of network hyperparameters and the training process. Section \ref{sec:experiments} experimentally validates the proposed control model using a legged robot. Section \ref{sec:conclusion} summarizes the main contributions of this paper and outlines future research directions.

\section{Related Work}
\label{sec:relatedwork}
In recent years, RL has made significant progress in the field of bipedal robot locomotion control, particularly forming two mainstream research directions in this area. One of these focuses on traversing complex terrain. To overcome the issue of robots being stuck in place on complex terrain, researchers have adopted a phased training strategy: initially, preset motion functions are used to enable the robot to master basic walking abilities on flat ground, followed by the introduction of curriculum learning to gradually increase terrain complexity \cite{20}. In terms of implementation, the locomotion control task is decomposed into two sub-goals: tracking of foot contact points and stabilization of the upper body posture \cite{21}. At the same time, a strategy for lifting the feet is trained by adding fractal noise to the terrain height field, effectively enhancing the robot's parkour capabilities on complex terrain \cite{22}. Addressing the temporal dependence of bipedal motion information, the latest research employs an input strategy that integrates current and historical observations, introduces a causal Transformer architecture to capture temporal features using self-attention mechanisms, and combines teacher-student strategies for terrain knowledge distillation, ultimately achieving robust locomotion control on complex terrain \cite{23}.

Human-likeness of motion is another important research branch in this field. Among these, reinforcement learning methods based on reference trajectories achieve human-like motion by using reward functions to continuously approximate remapped motion capture data \cite{24}. However, the approximation of whole-body joint angles based on RL faces certain challenges in physical deployment. Therefore, some studies propose constraining the upper body motion with motion capture data while releasing the constraints on the lower body (two legs) to enable more flexible tracking of commands \cite{25}. To ensure that motion capture data can be effectively mapped onto different types of robots, some studies have collected data from certain robots and used supervised learning methods to jointly train the robot data with the corresponding motion capture data, thereby generating motion data suitable for other types of robots \cite{26}. In addition, imitation learning methods based on generative adversarial networks (GANs) make the robot's motion style highly similar to that of the motion capture data through GAN networks, further enhancing the human-likeness of motion \cite{27}. However, pure imitation learning cannot surpass the performance of the data itself. Therefore, studying the internal mechanisms of the data becomes particularly important. Considering that the motion of bipedal robots has a certain periodicity, the Fourier Latent Dynamics (FLD) method enhances the interpolation and generalization capabilities of motion learning algorithms by mining the spatiotemporal structure of the motion space, solving the problem of sparse reference trajectory data, and enabling accurate state transition predictions on unseen motions \cite{11}. To further improve the performance of FLD, some studies have applied mean squared error (MSE) constraints to the latent embeddings of the input data, effectively preventing the decoder from generating meaningless actions and thereby further optimizing the robot's motion strategy \cite{28}.

\section{Method}
\label{sec:method}

Our goal is to enable the TRON1 robot to autonomously traverse various complex terrain obstacles without relying on terrain information and without undergoing teacher-student staged training.

\begin{figure*}[t]
    \vspace*{2mm}
    \centering
    \includegraphics[width=1\linewidth]{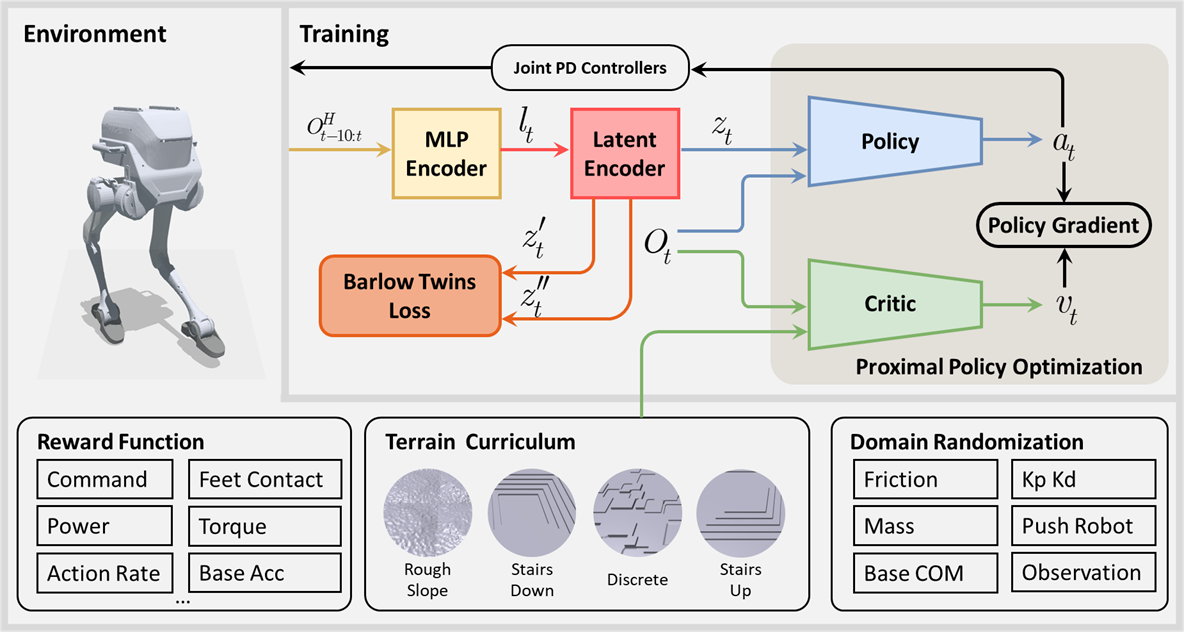}
    \caption{The overview of the proposed BarlowWalk. Within an asymmetric actor-critic framework, the policy and critic are trained using PPO. Specifically, the policy takes as input the observations excluding linear velocity, as well as the latent representations of historical observations from two sets of encoders. In contrast, the critic receives observations and privileged terrain information. Moreover, by applying the Barlow Twins loss to the latent representations over consecutive time steps, the latent features can be further optimized for better representation.}
    \label{fig:overview}
\end{figure*}

\subsection{Hardware Platform}
In our research, we utilized the footboard mode of the TRON1 \cite{29} bipedal robot from Limx Dynamics. The robot stands at 845 mm in height and weighs less than 20 kilograms. It is powered by electric motors in its joints. The robot features a total of eight degrees of freedom, with four degrees of freedom in each leg. The appearance and joint structure of the robot are shown in Fig.\ref{fig:robot}.

\subsection{Overview}
The overview of BarlowWalk is shown in Fig.\ref{fig:overview}. Specifically, our training is end-to-end and can be completed in just one stage, with different inputs for the policy and the critic.

\subsection{Barlow Twins}

\begin{figure}[t]
    \centering
    \includegraphics[width=1\linewidth]{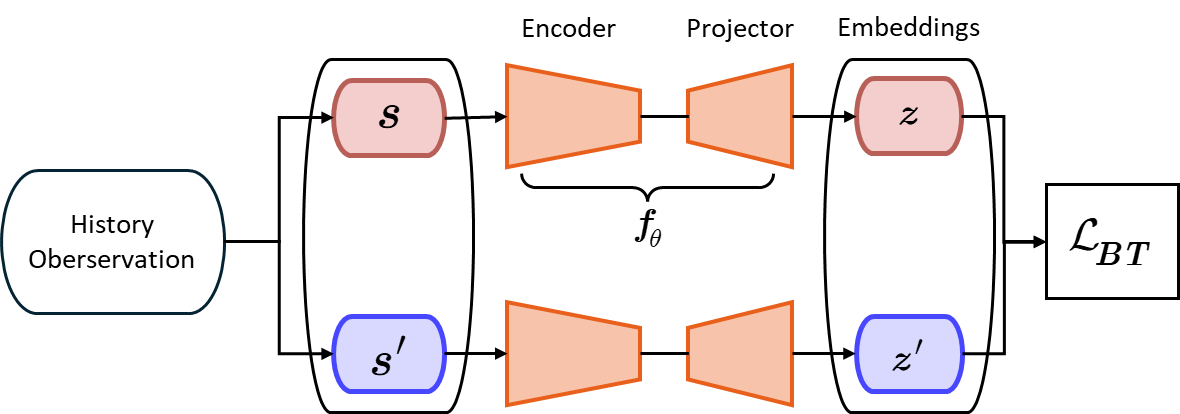}
    \caption{Schematic diagram of Barlow Twins.}
    \label{fig:bt}
\end{figure}

Barlow Twins\cite{add1} is a recently proposed advanced self-supervised learning method. Its core idea is to minimize the redundancy between representations of the same input under different data augmentations, thereby learning features that are both discriminative and generalizable. Compared with traditional contrastive learning methods, Barlow Twins introduces a redundancy reduction term, which eliminates the need for negative samples and offers higher stability and better scalability. The framework diagram of BARLOW TWINS is shown in Fig.\ref{fig:bt}.

\subsection{Training Pipline}
In the training process of TRON1, it can be described as a POMDP, represented by the tuple \((\mathcal{S}, \mathcal{O}, \mathcal{A}, \mathcal{R}, p, \gamma)\). Here, \(\mathcal{S}\) is the set of all states in the environment, \(\mathcal{O}\) is the set of observable state spaces, containing only the environmental information observable by the agent, \(\mathcal{A}\) is the set of all actions that the agent can perform, \(\mathcal{R}\) assigns a reward to each state-action pair, \(p\) is the probability distribution of state transitions after executing an action. \(\gamma \in [0, 1]\) is used to balance the influence of future rewards on current decisions. At each time step \(t\), the agent receives an environmental observation \(o_t \in \mathcal{O}\), follows a policy \(\pi(a_t | o_t)\) to choose an action \(a_t \in \mathcal{A}\). The environment transitions to a new state \(s_{t+1} \sim p(s_{t+1} | s_t, a_t)\), and the agent receives a reward \(r_t = \mathcal{R}(s_t, a_t)\). The ultimate goal is to maximize the cumulative reward to optimize the strategy:

\begin{equation}
J(\theta) = \mathbb{E}_{\pi_{\theta}} \left[ \sum_{t=0}^{T-1} \gamma^t r_t \right]    
\end{equation}

For traditional PPO, stability and convergence are ensured by limiting the magnitude of policy updates. During each policy update, PPO constrains the change in policy by introducing a clipping function, thereby balancing exploration with convergence. The optimization objective can be expressed in the following form:

{\small
\begin{equation}
\begin{aligned}
\mathcal{L}_{\pi} = \min \Bigg[ & \frac{\pi(a_t | o_t)}{\pi_b(a_t | o_t)} A^{\pi_b}(o_t, a_t),  \\
& \text{clip} \left( \frac{\pi(a_t | o_t)}{\pi_b(a_t | o_t)}, 1 - \varepsilon, 1 + \varepsilon \right) A^{\pi_b}(o_t, a_t) \Bigg]
\end{aligned}
\end{equation}
}

Here, \( \pi \) is the target policy to be optimized, \( \pi_b \) is the behavior policy used for data collection, and \( \varepsilon \) represents the clipping range of PPO.

In this study, we introduce self-supervised learning techniques to encode past time-step observations, with specific operations as follows. Since the observations include linear velocity, which cannot be directly measured in actual display scenarios, the historical observations input to the Multilayer Perceptron (MLP) encoder do not contain linear velocity information. We first remove the historical observations of the past 10 time steps that include linear velocity, denoted as \texttt{obs\_old\_hist}. Subsequently, we update these data with the latest observations to obtain \texttt{obs\_new\_hist}. Then, we use the last 5 time steps of \texttt{obs\_new\_hist} ( \texttt{obs\_new\_hist[5:10]}) as the input to the MLP encoder. The results are as follows:

\begin{table}[t] 
    \vspace{0.3cm} 
    \caption{Observations and Their Properties}
    \label{table1}
    \centering
    \begin{tabular}{lcc}
        \toprule
        \textbf{Observation} & \textbf{Dim.} & \textbf{Noise Range} \\
        \midrule
        Body velocity  $\mathbf{v}_b$ & $3$ & $0.1$ \\
        Body angular velocity $\boldsymbol{\omega}_b$ & $3$ & $0.2$ \\
        Body gravity $\hat{\mathbf{g}}$ & $3$ & $0.05$ \\
        Commanded velocities $\begin{Bmatrix} c_x \text{ (forward)} \\ c_y \text{ (lateral)} \\ c_{\omega} \text{ (yaw)} \end{Bmatrix}$ & $3$ & $0$ \\
        Joint positions $\mathbf{q}$ & $8$ & $0.01$ \\
        Joint velocities $\dot{\mathbf{q}}$ & $8$ & $1.5$ \\
        Action $\mathbf{a}$ & $8$ & $0.01$ \\
        Phase $\begin{Bmatrix} \sin(\phi) \\ \cos(\phi)  \end{Bmatrix}$ & $2$ & $0$ \\
        \bottomrule
    \end{tabular}
\end{table}

\begin{table}[t]
    \centering
    \caption{Network Input and Output Dimensions}
        \label{table2}
    \begin{tabular}{lcc}
        \toprule
        \textbf{Component} & \textbf{Input Dimension} & \textbf{Output Dimension} \\
        \midrule
        MLP Encoder & $175$ & $64$ \\
        Latent Encoder & $64$ & $16$ \\
        Policy & $51$ & $8$ \\
        Barlow Encoder & $16$ & $64$ \\
        Critic & $225$ & $1$ \\
        \bottomrule
    \end{tabular}
\end{table}

\begin{equation}
    l_t = \text{MLP\_enc}(\text{obs\_new\_hist}[5{:}10])
\end{equation}

Then, the output of the MLP encoder is used as the input for the latent encoder, calculated as follows:

\begin{equation}
    z_t = \text{latent\_enc}(l_t)
\end{equation}

Finally, both \( z_t \) and the current observation are fed into the policy network for training.

During this period, Barlow Twins are utilized for self-supervised learning, with the goal of learning features that possess strong representational capabilities. Therefore, an additional Barlow encoder is required.
For each time step, \texttt{obs\_old\_hist[5:10]}) and \texttt{obs\_new\_hist[5:10]}) are fed into the MLP encoder to obtain:

\begin{equation}
    l''_t = \text{MLP\_enc}(\text{obs\_new\_hist}[5:10])
\end{equation}

\begin{equation}
    l'_t = \text{MLP\_enc}(\text{obs\_old\_hist}[5:10])
\end{equation}

Then, the outputs of MLP encoder are input into the latent encoder, yielding the following results:

\begin{equation}
    z''_t = \text{latent\_enc}(l''_t)
\end{equation}
\begin{equation}
    z'_t = \text{latent\_enc}(l'_t)
\end{equation}

The output features from the latent encoder are then fed as input to the Barlow encoder for feature refinement, with the computation process formulated as follows:

\begin{equation}
    u''_t = \text{Barlow\_enc}(z''_t)
\end{equation}
\begin{equation}
    u'_t = \text{Barlow\_enc}(z'_t)
\end{equation}

Next, we calculate the cross-correlation matrix of the outputs from the latent encoder:

\begin{equation}
    C_{ij} = \frac{\sum_{b=1}^{N} u'_{t,b,i} u''_{t,b,j}}{\sqrt{\sum_{b=1}^{N} (u'_{t,b,i})^2} \sqrt{\sum_{b=1}^{N} (u''_{t,b,j})^2}} 
\end{equation}

\begin{table}[t]
    \vspace{0.3cm} 
    \centering
    \caption{Hyperparameters for Training}
    \label{table3}
    \begin{tabular}{lc}
        \toprule
        \textbf{PPO} & \textbf{Value}\\
        \midrule
        Batch size & $4096\times24$ \\
        Mini-batch size &  $4096\times6$ \\
        Number of epochs & $5$ \\
        Clip range & $0.2$ \\
        Entropy coefficient & $0.01$ \\
        Discount factor & $0.99$ \\
        GAE discount factor & $0.95$ \\
        Desired KL-divergence & $0.01$ \\
        Learning rate & adaptive \\
        Barlow $\lambda$ &  $5e^{-3}$ \\
        \bottomrule
    \end{tabular}
\end{table}

\begin{table*}[ht]
\vspace{0.3cm} 
\centering
\caption{Reward Terms for Tracking Rewards, Action Rewards, and Constraint Rewards}
\label{table4}
\begin{tabular}{llcc}
\toprule
\textbf{Classification} & \textbf{Reward} & \textbf{Equation} & \textbf{Weight} \\
\midrule
\multirow{7}{*}{Tracking rewards}
        & Linear velocity tracking & $\exp\left\{-\| \mathbf{v}^{\text{cmd}}_{xy} - \mathbf{v}_{xy} \|_2^2 / 0.25\right\}$ & $5.0$ \\
        & Angular velocity tracking & $\exp\left\{-\| \boldsymbol{\omega}^{\text{cmd}}_{\text{yaw}} - \boldsymbol{\omega}_{\text{yaw}} \|_2^2 / 0.25\right\}$ & $2.5$ \\
        & Action smoothness & $\| \mathbf{a}_t - 2 \mathbf{a}_{t-1} - \mathbf{a}_{t-2} \|_2^2$ & $-0.01$ \\
        & Angular velocity penalty & $\| \boldsymbol{\omega}_{xy} \|_2^2$ & $-0.05$ \\
        & Body height penalty & $\| \hat{\mathbf{h}}^{\text{cmd}} - \mathbf{h} \|_2^2$ & $-10.0$ \\
        & Orientation & $g_x^2 + g_y^2$ & $-1.0$ \\
        & Feet clearance & $\left( h^{\text{target}}_{\text{foot}} - h_{\text{foot}} \right)^2 \cdot v_{xy}$ & $1.0$ \\
\midrule
\multirow{5}{*}{Action rewards}
        & Torques & $\| \boldsymbol{\tau} \|_2^2$ & $-8e-5$ \\
        & Powers & $\| \mathbf{tq} \|^2$ & $-2e-3$ \\
        & Degrees of freedom velocity & $-\| \dot{\mathbf{q}} \|_2^2$ & $-1e-3$ \\
        & Degrees of freedom acceleration & $-\| \ddot{\mathbf{q}} \|_2^2$ & $-2.5e-7$ \\
        & Feet swing height & $\left( h_{\text{foot contact}} - 0.1 \right)^2$ & $-20.0$ \\
\midrule
\multirow{5}{*}{Constraint rewards}
        & Contact & $\left( \phi_{\text{foot}} < 0.55 \right) \oplus \left( F_{\text{foot,z}} > 1 \right)$ & $0.18$ \\
        & Base acceleration & $\exp \left\{ -\| \boldsymbol{a}_{\text{root}} \|_3 \right\}$ & $0.2$ \\
        &Feet contact forces & $\text{max}(0, \text{max\_contact\_force} - \| \mathbf{F}_{\text{contact}} \|) \cdot \mathbb{I}(\| \mathbf{F}_{\text{contact}} \| > 1)$ & $-0.002$\\
        &Feet air time & $(\text{feet\_air\_time} - 0.5) \cdot \mathbb{I}(\text{first\_contact}) \cdot \mathbb{I}(\| \mathbf{v}_{\text{cmd}} \| > 0.1)$ & $1.0$ \\
        & Feet contact number & $\left[ \left( \phi_{\text{foot}} < 0.55 \right) \oplus \left( F_{\text{foot,z}} > 5 \right) - 0.3 \right] / 2$ & $1.2$ \\
\bottomrule
\end{tabular}
\end{table*}

The cross-correlation matrix \( C \) is calculated from the outputs of the latent encoder. Each element \( C_{ij} \) in the matrix represents the correlation between the \( i \)-th and \( j \)-th features across all samples in a batch. The matrix \( C \) has dimensions \( d \times d \), where \( d \) is the dimensionality of the features. \( u'_{t,b,i} \) represents the \( i \)-th feature of the \( b \)-th sample from the past history at time \( t \). Similarly, \( u''_{t,b,j} \) represents the \( j \)-th feature of the \( b \)-th sample from the current history at the same time \( t \). \( N \) denotes the batch size, which is the total number of samples in the batch. The dimensionality of each feature is represented by \( d \).Thus, a new loss function needs to be added:

\begin{equation}
    \mathcal{L}_{BT} = \sum_{i=1}^{d} (C_{ii} - 1)^2 + \lambda \sum_{i \neq j} C_{ij}^2
\end{equation}

Here, \( C_{ii} \) represents the autocorrelation, while \( C_{ij} \) represents the cross-correlation. The final optimization objective, namely the loss function, is as follows:

\begin{equation}
    \mathcal{L}_{\text{total}} = \mathcal{L}_{\pi} + \mathcal{L}_{BT}
\end{equation}

\subsection{State and Action}

In the foot - plate mode of TRON1, the dimension of the action corresponds to two legs, i.e., $a_t \in \mathbb{R}^8$. Regarding the State, the specific state dimensions are shown in TABLE \ref{table1}. Therefore, for the policy, the observation input to the MLP encoder is $O_{\text{MLP\_enc}} \in \mathbb{R}^{175}$; while the observation input to the critic, which includes terrain information, is $O_{\text{critic}} \in \mathbb{R}^{225}$. The bolded part in TABLE \ref{table1} represents the parts that are exclusive to the critic.

\subsection{Policy Architecture and Parameter }
In this study, the core networks of both the actor and the critic employ Gated Recurrent Units (GRUs). Each GRU has only one layer with a hidden size of 64. A MLP follows the GRU, and the dimension of this MLP is 32. Additionally, there are three encoders: the MLP encoder, the Latent encoder, and the Barlow encoder. Their network dimensions are [128, 64], [32, 16], and [16, 64], respectively. For easy reference of the last - dimensional information of the input and output of each network, the details are presented in TABLE \ref{table2}. Detailed training hyperparameters are provided in TABLE \ref{table3}.

To achieve robust locomotion across various terrains, implementing an appropriate training curriculum strategy is of utmost importance. We have adopted a terrain curriculum and an instruction curriculum, and selected four terrain types for training: slopes, rough slopes, stairs, and discrete obstacles. Details are shown in Fig. \ref{fig:overview}. 

\subsection{Reward}
The reward function is formulated as a multi-objective optimization problem, combining three distinct components:
\begin{enumerate}
    \item \textbf{Tracking Performance Reward }\( r_{\text{tracking}} \): Measures the accuracy of tracking desired base velocity, height, and posture commands.
    \item \textbf{Action Smoothness Penalty }\( r_{\text{action}} \): Discourages excessive joint torque fluctuations and minimizes energy expenditure.
    \item \textbf{Constraint Satisfaction Reward }\( r_{\text{constraint}} \): Enforces safety-critical conditions, including contact force limits and joint motion ranges.
\end{enumerate}

The total reward \( r_{\text{total}} \) is computed as a weighted sum of these components:

\begin{equation}
    r_{\text{total}} = r_{\text{tracking}} + r_{\text{action}} + r_{\text{constraint}}
\end{equation}

The specific settings of the reward function are shown in TABLE \ref{table4}.

\begin{table}[t]
\caption{Randomization Terms}
\label{table5}
\begin{center}
\begin{tabular}{lll}
\toprule
\textbf{Randomization Term} & \textbf{Range} & \textbf{Unit} \\
\midrule
Link mass & $[0.8, 1.2] \times \text{nominal value}$ & \text{Kg} \\
Payload mass & $[-1, 3]$ & \text{Kg} \\
CoM of base & $[-7.5, 7.5] \times [-5, 5] \times [-5, 5]$ & \text{cm} \\
Friction & $[0.2, 1.25]$ & - \\
Restitution & $[0, 1]$ & - \\
Joint $K_p$ & $[0.9, 1.1] \times \text{nominal value}$ & \text{N-rad} \\
Joint $K_d$ & $[0.9, 1.1] \times \text{nominal value}$ & \text{N-rad/s} \\
Motor strength $K_d$ & $[0.8, 1.2] \times \text{nominal value}$ & \text{N} \\
\bottomrule
\end{tabular}
\end{center}
\end{table}

\subsection{Domain Randomization}
Domain randomization is a key technique in RL for enhancing policy transferability. By systematically perturbing simulation environment parameters during training, it effectively bridges the simulation-to-reality gap. In this study, we implement randomization on both robot dynamics and environmental interaction parameters. For robot dynamics, we randomize the center-of-mass position and joint motor power. For environmental interactions, we dynamically adjust ground friction coefficients and apply randomized external forces to the robot's torso. All parameters are uniformly sampled within predefined ranges, with specific values detailed in TABLE \ref{table5}. This approach significantly improves the policy's adaptability to variations in dynamic parameters.

\begin{figure}
    \vspace*{2mm} 
    \centering
    \includegraphics[width=1\linewidth]{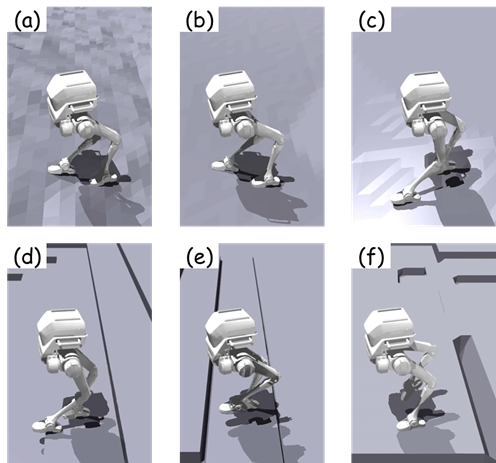}
    \caption{This study develops six typical terrain environments with adjustable difficulty levels. Through systematic training across all terrains, the robot demonstrates effective adaptation to various complex terrains encountered in human activity scenarios.(a) Rough Ground; (b) Slope Up; (c) Slope Down; (d) Stairs Down; (e) Stairs Up; (f) Obstacles.}
    \label{fig:simulate}
\end{figure}

\subsection{Curriculum Learning}
To tackle the motion stability challenges of TRON1 robots in unstructured terrains, this study proposes a phased progressive terrain training strategy. By integrating multimodal terrain generation technology with dynamic difficulty adjustment algorithms, this approach significantly mitigates common issues in conventional reinforcement learning such as inadequate exploration efficiency and policy instability in complex environments. Building on the correlation between motor skill generalization and terrain diversity, the research team developed a training platform incorporating six characteristic terrain scenarios: irregular rough surfaces, bidirectional slopes (ascent/descent), multi-level staircases (up/down), and randomly distributed obstacles as core training elements.

To address the policy instability during the initial training phase of reinforcement learning and the low convergence efficiency caused by direct training on complex terrains, this study  designs ten standardized straight paths as benchmark training environments. The simulation system is constructed based on high-precision 20×10 grid height field maps, enabling accurate simulation of diverse terrains through parametric modeling.

\section{Experiments}
\label{sec:experiments}
Using IsaacGym \cite{30}, we trained 4,096 parallel TRON1 agents on an RTX 4090 GPU across diverse terrains. Each 20-second episode (50Hz control) terminated upon robot fall or timeout. PD control parameters: hip (kp=100,kd=1.5), knee (kp=150,kd=1.5), ankle (kp=45,kd=0.8). Policy updates occurred every 24 steps, with complete training (4 hours) showing continuous improvement. Test scenarios in Fig. \ref{fig:real}.

\begin{figure}
    \vspace*{2mm}
    \centering
    \includegraphics[width=1\linewidth]{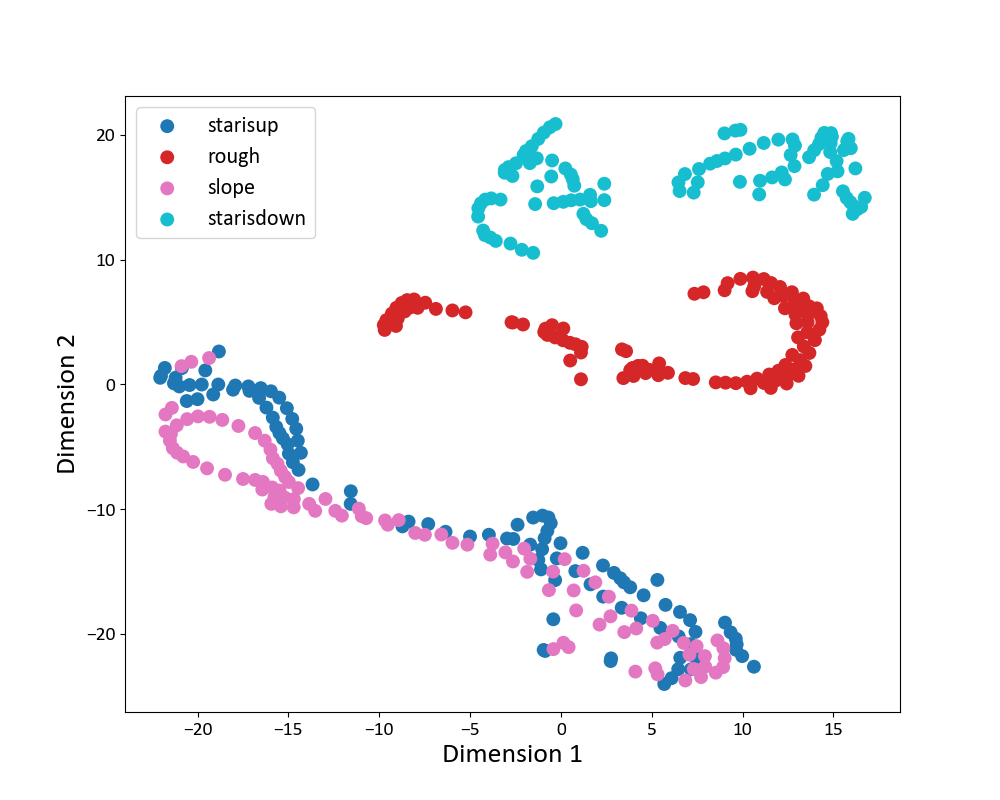}
    \caption{t-SNE visualization of terrain.}
    \label{fig:tsne}
\end{figure}

\begin{figure}
    \centering
    \includegraphics[width=1\linewidth]{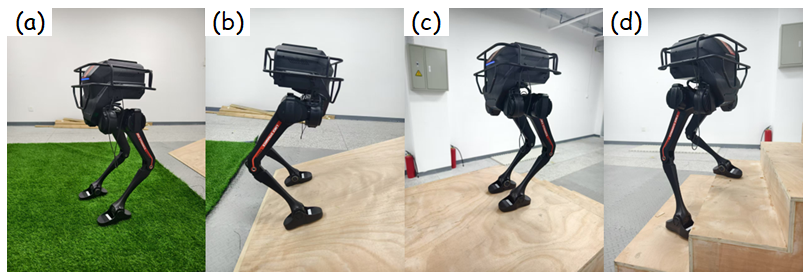}
    \caption{Robot hardware test scenarios:(a) TRON1 navigating grassy terrain, (b) slope climbing demonstration, (c) platform obstacle negotiation, (d) stair traversal capability verification.}
    \label{fig:real}
\end{figure}

\subsection{Experimental Results}
Six types of terrains were designed, encompassing rough flat surfaces, uphill slopes, downhill slopes, ascending stairs, descending stairs, and obstacle terrains,  as shown in Fig.\ref{fig:simulate}. Experimental results demonstrate that TRON1 achieves robust locomotion across all six terrain types. Furthermore, within the predefined 10-level terrain difficulty assessment system, the robot consistently passes challenges up to difficulty level 6.

\begin{figure}
    \vspace*{2mm}
    \centering
    \includegraphics[width=1\linewidth]{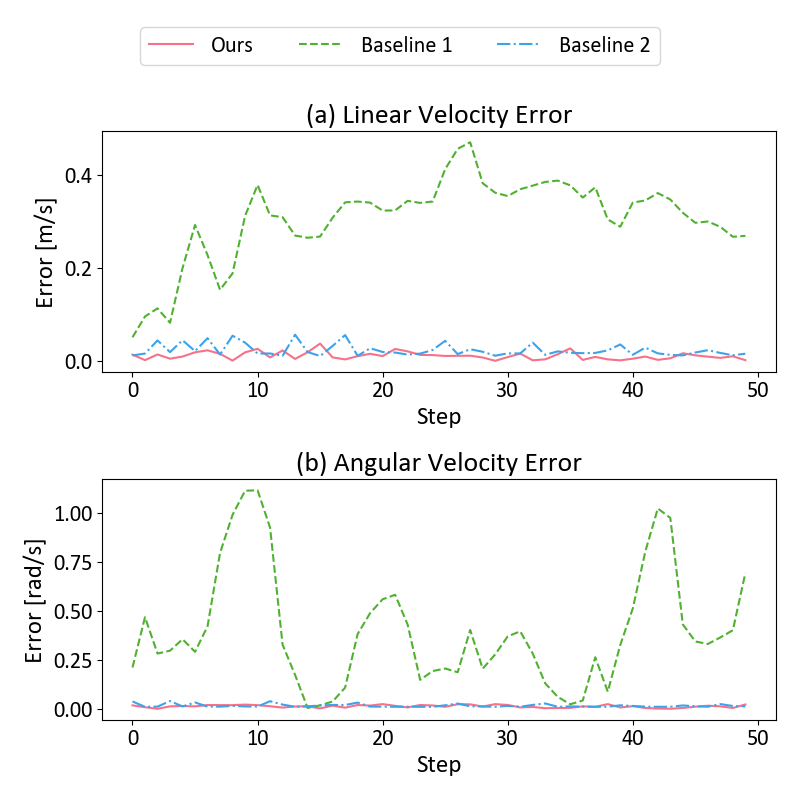}
    \caption{(a): tracking errors in linear velocity during the initial 50 walking steps, (b): tracking errors in angular velocity during the initial 50 walking steps. }
    \label{fig:error}
\end{figure}

\subsection{Latent Space Analysis}
To evaluate the latent representations in the feature space, we analyzed the latent features output by the latent encoder while the robot traversed various terrains, collecting data corresponding to its motion on each terrain. Using the t-distributed stochastic neighbor embedding (t-SNE) method, we visualized the latent representations of the robot across different terrains. Since the obstacle terrain contained information from other terrains, it was excluded from this analysis, and we focused only on the distribution of the remaining terrains, as shown in Figure 4. The results revealed distinct distribution patterns of the embedded latent representations for different terrain behaviors. Notably, the latent representations corresponding to uphill slopes and ascending stairs exhibited significant similarity, indicating that the robot perceives climbing an incline and ascending stairs as similar processes.

\subsection{Comparative Experiment}
To demonstrate the superiority of the proposed method, we conduct comparative experiments with two baseline approaches, focusing on tracking errors of TRON1's linear and angular velocities. The experimental configuration includes three methods:

\begin{itemize}
    \item \textbf{Ours (BarlowWalk):} The complete proposed framework;
    \item \textbf{Baseline 1 (SLR\cite{19})}: Adapting the SLR framework by replacing its quadruped robot with TRON1 while maintaining reward function settings analogous to our method; 
    \item \textbf{Baseline 2:} A degraded version eliminating all terrain perception, where 5 - step observation windows are fed solely into an MLP encoder (for both policy and critic networks).
\end{itemize}

The experimental results are presented in Fig.\ref{fig:error}. As can be observed from  Fig.\ref{fig:error} (a), our method demonstrates slight improvements over Baseline 1, exhibiting lower tracking errors in linear velocity. This indicates that the proposed self-supervised approach can effectively extract useful latent features. In contrast, Baseline 2, which utilizes only a historical proprioceptive encoder, shows significantly poorer velocity tracking performance. From  Fig. \ref{fig:error} (b), it can be seen that our method achieves comparable angular velocity tracking errors to Baseline 1, while Baseline 2 continues to demonstrate inferior performance.

\subsection{Ablation Study for the Learning Framework}

\begin{figure}
    \vspace*{2mm}
    \centering
    \includegraphics[width=1\linewidth]{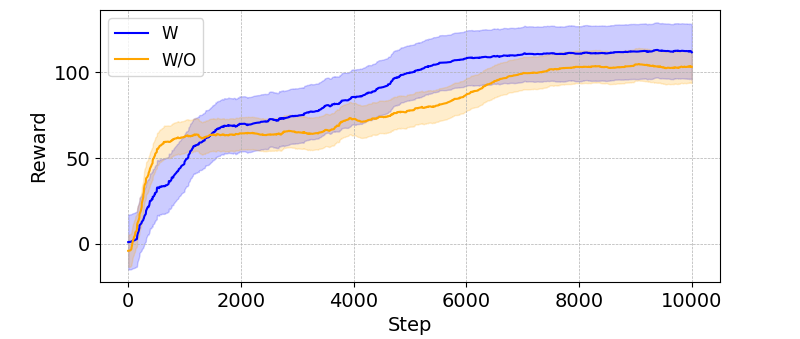}
    \caption{Comparison of mean reward functions.}
    \label{fig:reward}
\end{figure}

This study introduces a new loss function term based on the original PPO algorithm's loss function. To evaluate the effectiveness of this improvement, we designed systematic ablation experiments, conducting comparative analyses primarily along the following two dimensions:

\begin{figure}
    \centering
    \includegraphics[width=1\linewidth]{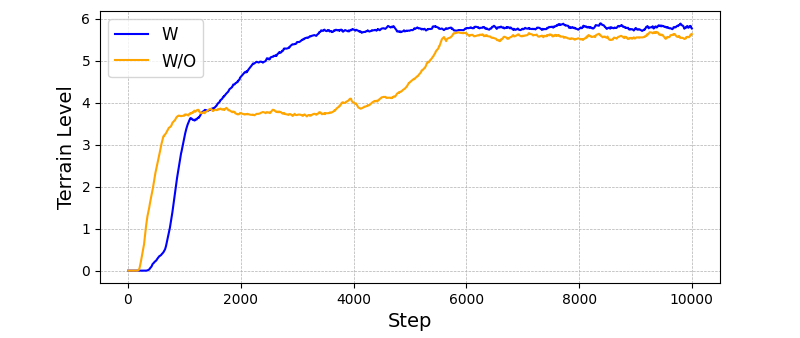}
    \caption{Comparison of Terrain level.}
    \label{fig:terrain}
\end{figure}

\begin{itemize}  
    \item \textbf{Algorithm convergence comparison}: Assessing performance differences through the variation curves of the reward function during the training process;  
    \item \textbf{Generalization capability comparison}: Under the same number of training iterations, comparing the algorithms' adaptive performance in terrains of varying difficulty levels.  
\end{itemize}

As shown in Fig. \ref{fig:reward}, under the same number of training steps, the model without the Barlow Twins loss term exhibits faster reward growth in the early training phase. However, in the mid-to-late stages, the model with the Barlow Twins loss term demonstrates accelerated reward improvement and more stable convergence, indicating its superior optimization efficiency.

Fig. \ref{fig:terrain} further supports this observation: the model without Barlow Twins initially progresses slowly in terrain level adaptation, but after the early phase, it rapidly advances and reaches near Level 6 at around 4,000 steps, ultimately achieving stable convergence.

\section{Conclusion}
\label{sec:conclusion}

This study proposes an innovative bipedal robot motion control framework, whose core innovation lies in abandoning traditional knowledge distillation techniques in favor of a self-supervised feature learning approach based on proprioceptive data. The framework maps current and historical proprioceptive data to a latent feature space and employs the Barlow Twins algorithm for self-supervised learning of continuous-time latent features, thereby enabling the efficient extraction of high-quality motion features from sparse data. In terms of control strategy, we designed a multi-objective reward function system to effectively guide the robot's adaptive walking behavior in complex terrains. Considering the importance of continuous-time sequence information, we used GRU network architecture to model temporal dependencies. To verify the performance of the framework, we conducted systematic comparative experiments, including motion tests under various terrain conditions. The experimental results demonstrate that the control framework can achieve robust motion with relatively high precision in complex terrains. Looking to the future, we plan to further optimize this framework to perform more complex motion control tasks and ultimately deploy it to physical robot platforms for practical application verification.

\addtolength{\textheight}{-12cm}   




\bibliographystyle{IEEEtran} 
\bibliography{IEEEexample}

\end{document}